\pdfoutput=1
\documentclass[11pt]{article}

\usepackage[final]{acl}

\usepackage{xcolor}
\usepackage{times}
\usepackage{latexsym}
\usepackage{algorithm}
\usepackage{algpseudocode}
\usepackage{multirow}
\usepackage{tabularx}
\usepackage{afterpage}
\usepackage{amsmath}
\usepackage{booktabs}

\usepackage[most]{tcolorbox}

\usepackage[T1]{fontenc}
\usepackage[utf8]{inputenc}

\usepackage{microtype}

\usepackage{inconsolata}

\usepackage{graphicx}

\title{Auto prompting without training labels: An LLM cascade for product quality assessment in e-commerce catalogs}

\author{
 \textbf{Soham Satyadharma\textsuperscript{1}},
 \textbf{Fatemeh Sheikholeslami \textsuperscript{1}},
 \textbf{ Swati Kaul\textsuperscript{1}}, \\
 \textbf{ Aziz Umit Batur\textsuperscript{2}},
 \textbf{Suleiman A. Khan\textsuperscript{1}}
\\
 \textsuperscript{1} Amazon Catalog AI,
 \textsuperscript{2} formerly at Amazon Catalog AI
\\
 \small{
 {ssatyadh@amazon.com, shfateme@amazon.com, kauswati@amazon.com,  umitbatur@gmail.com, suleimkh@amazon.com}
 }
}

\begin{document}
\maketitle
\begin{abstract}
We introduce a novel, training free cascade for auto-prompting Large Language Models (LLMs) to assess product quality in e-commerce. Our system requires no training labels or model fine-tuning, instead automatically generating and refining prompts for evaluating attribute quality across tens of thousands of product category–attribute pairs. Starting from a seed of human-crafted prompts, the cascade progressively optimizes instructions to meet catalog-specific requirements. This approach bridges the gap between general language understanding and domain-specific knowledge at scale in complex industrial catalogs. Our extensive empirical evaluations shows the auto-prompt cascade improves precision and recall by 8–10\% over traditional chain-of-thought prompting. Notably, it achieves these gains while reducing domain expert effort from 5.1 hours to 3 minutes per attribute - a 99\% reduction. Additionally, the cascade generalizes effectively across five languages and multiple quality assessment tasks, consistently maintaining performance gains.

\end{abstract}

\section{Introduction}\label{sec:introduction}
%DO NOT EDIT/Suleiman
% Product catalogs are the cornerstone of e-commerce stores, acting as the critical interface between sellers and buyers where the accuracy of product information directly impacts user experience and business outcomes \cite{amsl2023presenting, lv2022impact, sadinle2022semi}. Each product is characterized by a comprehensive set of data fields referred commonly as attributes. These attributes fall in two broad categories namely, unstructured attributes (such as free-text titles, descriptions, and images) and structured attributes (including specific feature-value pairs like color, size, and material specifications) \cite{nikolakopoulos2023sage}. Despite advances in data management systems, maintaining consistency and accuracy across these diverse information fields remains a significant challenge. While Large Language Models (LLMs) have revolutionized natural language processing capabilities \cite{zubiaga2024natural, hadi2023large, min2023recent, huang2024leveraging, li2023large}, their application to specialized industrial tasks like product information quality assessment remains complex, primarily due to the requirement for precise, domain-specific strategies.

% Soham
Product catalogs are the cornerstone of e-commerce, where the accuracy of product information directly impacts user experience and business outcomes \cite{amsl2023presenting, lv2022impact, sadinle2022semi}. Each product is defined by unstructured attributes (free-text like titles or descriptions) and structured attributes (feature-value pairs like color or size) \cite{nikolakopoulos2023sage}. A fundamental challenge in maintaining catalog quality is ensuring the alignment between these two attribute types, as inconsistencies frequently arise from discrepancies between seller descriptions and how attributes are formally modeled \cite{schmidts2020catalog}. This complexity is twofold: First, inherent semantic ambiguities make verification difficult. For example, identifying the base material of a walking stick from the description, \textit{wood construction with a steel spike and rubber tip}, requires disambiguating material of each component. Second, attribute values  often rely on implicit sequential reasoning, such as inferring that a product talks about pet food or human food, and using this inference to predict if valid age is \textit{puppies} or \textit{young adult}. For additional examples see:  \S\ref{sec:qualitative_results}. 

%Second, attribute values often rely on implicit context, such as inferring an age range for pet food when a listing only mentions \textit{puppies} without specifying an explicit age. We explain these examples in detail in \S\ref{sec:qualitative_results}. 

These alignments and nuances vary for each product category-structured attribute (PC-SA) pair, which exhibits distinct characteristics and semantic interpretations. Here PC is a group of similar products (e.g., speakers or shirts), and SA is a structured attribute (e.g., color or material).
While Large Language Models (LLMs) offer strong reasoning capabilities \cite{zubiaga2024natural, hadi2023large, min2023recent, huang2024leveraging, li2023large}, steering them for such a specialized task with ten's of thousands of implicit nuances remains a complex challenge. This is further compounded by misalignment of LLM's general knowledge with specialized terminology and quality expectations of e-commerce stores.
Single, general-purpose solutions (zero-shot or few-shot) struggle to effectively capture these variations \cite{jiang2024hallucination}. A promising direction is to steer the LLM to determine correct values through domain-aware, case-specific instructions for each PC-SA pair. This is a manual task suitable for subject matter experts and estimated to require over 3,000 human-days for over 12,000 PC-SA pairs.

%The challenge of identifying these alignments is magnified significantly at the scale of tens of thousands of product category-structured attribute (PC-SA) pairs, where a PC is a group of similar products (e.g., PC1: speakers and PC2: shirts), and each SA is a structured attribute (e.g., SA1: color, SA2: material). Each pair exhibits distinct characteristics and semantic interpretations, making it computationally intractable for a single, general-purpose zero-shot solution to effectively capture these variations. While Large Language Models (LLMs) offer strong reasoning capabilities \cite{zubiaga2024natural, hadi2023large, min2023recent, huang2024leveraging, li2023large}, their use in specialized tasks remain complex. Steering an LLM to determine correct values requires domain-aware, case-specific instructions for each PC-SA pair - a manual task estimated to require over 3,000 human-days -a challenge compounded by the misalignment between LLM's general knowledge, specialized terminology, and quality expectations of e-commerce stores.

\begin{figure}[t]
    \centering
    \includegraphics[width=0.95\columnwidth]{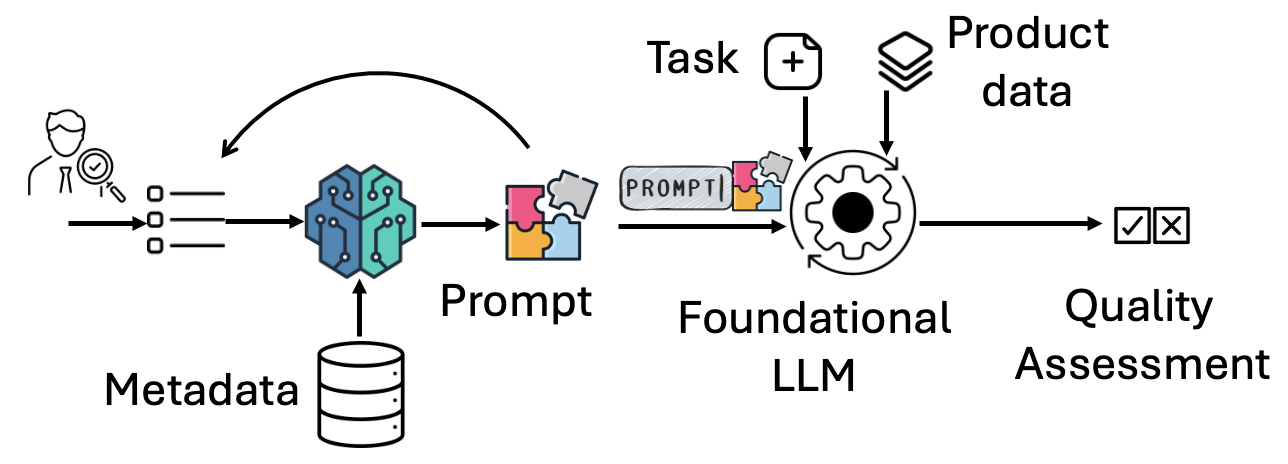}
    \vspace{-.1cm}
    \caption{ Auto-prompt LLM cascade for quality measurement. The cascade takes metadata and manually created few-shot examples to iteratively generate prompts, subsequently used to classify the attribute value. }
    \label{fig:aqumen}
\vspace{-.6cm}

\end{figure}

Previous attempts to address these challenges have shown promise but faced limitations. For instance, MetaBridge \cite{wang2020automatic} combines meta-learning with latent variable modeling to verify attributes but requires an exceedingly large number of labeled training samples for each PC-SA, rendering it practically infeasible at the scale of tens of thousands of PC-SA. On the other hand, zero-shot Chain-of-Thought (CoT) approaches have helped in simpler tasks \cite{wei2022chain, kojima2022large, zhang2022automatic, wang2022self}, but they lack the granularity to handle the thousands of heterogeneous relationships between PCs and SAs, which are needed to ensure domain-specific terminology, contextual nuances, and the hierarchical nature of product taxonomies. %These challenges underscore the need for more specialized methods that can adapt to category-specific  patterns while maintaining computational efficiency at the modern e-commerce store scale.

% To address these challenges, we propose an innovative LLM cascade approach (Figure \ref{fig:aqumen}) for auto-prompting language models to assess product quality across two catalog quality tasks: correctness and applicability (detailed in \S \ref{sec:datasets}). Our approach automates the generation and refinement of prompts through an iterative formulation that progressively builds and optimizes the prompts to steer off-the-shelf \emph{Foundational LLM} in performing nuanced quality assessment at scale. A core component of the prompts are PC-SA instructions which steer the foundational LLM in making product quality assessments. In the remainder of the paper, we use the terms prompt and instruction interchangeably. Starting with a seed of perfectly human-crafted PC-SA instructions, the cascade creates instructions for an SA across multiple product categories, enabling prompts to capture salient nuances across the diverse product types. For example, this generates prompts for SA material while specifying nuances for various product categories like sofas, televisions, shirts and, screws. The next iterations build upon these automatically generated instructions and refine them for target PC-SAs while incorporating detailed metadata such as PC and SA definitions.

To address these challenges, we introduce an innovative LLM cascade for large scale product quality assessment without any training labels or model fine-tuning (Fig. \ref{fig:aqumen}). Our approach steers off-the-shelf LLMs by iteratively generating and refining tens of thousands of prompts for two catalog quality tasks: correctness and applicability (\S\ref{sec:datasets}). A core component of the prompts are tailored PC-SA instructions which steer the LLM in making product quality assessments. Bootstrapping from a small set of human-authored examples, the cascade creates instructions for each SA across multiple PCs (e.g., material-sofas, material-screws). The cascade iteratively refines these instructions by incorporating detailed metadata, such as PC and SA definitions, valid-values etc. In the remainder of the paper, we use the terms prompt and instruction interchangeably.

The contributions of this paper are three fold. (1) We introduce a novel, training-free LLM cascade that automatically generates and refines prompts for e-commerce quality assessment, requiring no labeled data or model fine-tuning. (2) We demonstrate that our approach improves precision and recall by 8–10\% over the baseline across tens of thousands of diverse PC-SA pairs, while reducing human prompting effort by 99\%. (3) We establish that our method generalizes effectively across multiple quality tasks, foundational LLMs, and languages, all without the need for task or language specific training labels.

% This work represents a significant advancement in automated quality assessment for e-commerce stores, demonstrating how LLMs can be effectively aligned  through automated prompt engineering to perform complex, domain-specific tasks at scale, reducing human prompting efforts.

\section{Related work}
\textbf{Prompt engineering:} In-context learning through hand-crafted prompts have been shown to improve LLM performance. Approaches like Chain-of-Thought prompting \cite{wei2022chain,ma2023chain} and few-shot learning \cite{brown2020language,radford2019language} enhance LLM reasoning, though they rely on manual engineering and offer limiting scalability to write thousands of prompts. Our work builds on these by automating prompt generation at scale to create tens of thousands of domain-specific prompts while retaining benefits of structured prompt design offered by both.

\textbf{Automated prompt generation:} Automated prompt generation has shown promise in improving prompts, however, most methods focus on general-purpose tasks \cite{zhou2022large, opsahl2024optimizing} rather than domain-specific improvements. Approaches like reinforcement learning \cite{deng2022rlprompt} and evolutionary algorithms \cite{guo2023connecting} optimize prompts based on feedback; while clustering and failure-driven rules help select optimal prompts from synthetic candidates \cite{do2024automatic, gao2025prompt}. PRISM \cite{he2024automated} and PromptGen \cite{zhang2022promptgen} use iterative approaches for in-context learning. However, these techniques require labeled development sets or error feedback for optimization, whereas our method generates domain-specific prompts directly from knowledge hierarchies without such supervision.

Recently, hybrid approaches combining prompt engineering and fine-tuning have shown promising results, for example integrating prompt tuning with Bayesian regression \cite{wang2025sequential}, fine-tuning \cite{soylu2024fine} and reinforcement learning \cite{byun2024ares, kong2024prewrite}. However, unlike ours, these approaches require training labels for optimizing both models and prompts which is prohibitive at the scale of tens of thousands of PC-SA's.

\textbf{LLMs for e-commerce:} LLMs have shown benefit in several e-commerce applications such as, improving recommendation systems \cite{maragheh2023llm}, knowledge graph completion \cite{chen2023knowledge}, search \cite{rokon2024enhancement}, product discovery \cite{wang2024leveraging}, product matching \cite{herrero2024learning}, categorization \cite{cheng2024commerce, kathiriya2023optimizing}, and attribute extraction \cite{baumann2024using}. However, no prior work has investigated using LLMs for assessing product quality. 

\section{Method}
\label{headings}
\begin{figure}[t]
    \centering
    \includegraphics[width=0.95\columnwidth]{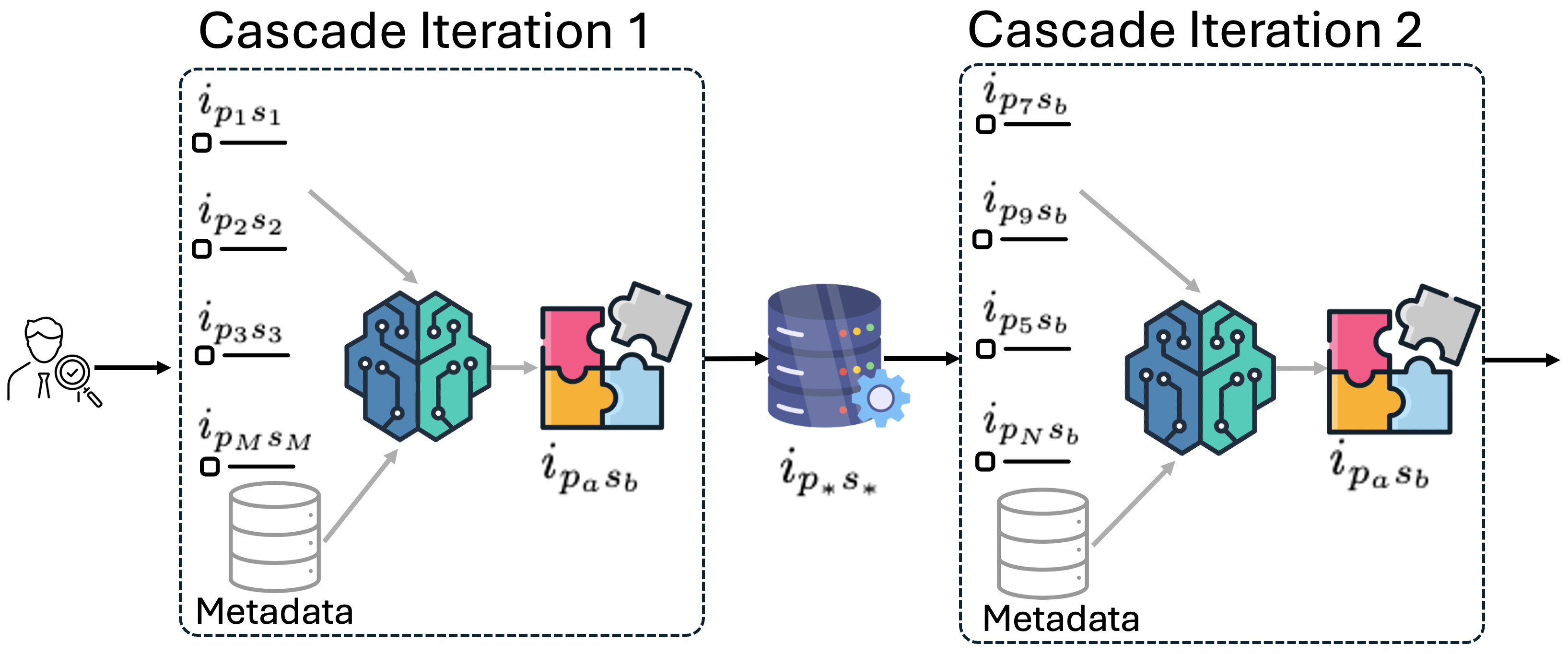}    
    \caption{Example of auto prompting cascade for instruction generation of PC denoted as $p_a$ and attribute  as $s_b$ across two iterations. Iteration 1 utilizes $M$ manually created global PC-SA instructions and is repeated for $M$ PC definitions to generate $M$ PC-SA instructions for the $s_b$ attribute that are used by iteration 2 to combine these examples with the definition for $p_a$ and $s_b$ to produce the PC-SA instruction for $p_a$ - $s_b$.}
    \label{fig:two_iteration_PA_instruction}
\end{figure}

\subsection{Problem formulation}\label{sec:problem_formulation}
Let $\mathcal{A}$ be the universe of products, where each product $a \in \mathcal{A}$ is characterized by a tuple  $(p_a, \{s_{a_i}\}_{i=1}^{|S|}, u_a)$, where $p_a \in \mathcal{P}$ is the PC, $s_{a_i} \in \mathcal{S}$ denotes the $i$-th structured attribute in the set of structured attributes (SAs) for product $a$, and $u_a \in \mathcal{U}$ denotes the unstructured attributes (UAs) for product $a$. By dropping subscript $a$ for simplicity, we define the task of quality assessment of SA $s_i$ for a given product $a$ as learning the classification function 
\begin{equation}
f: \mathcal{P} \times \mathcal{S} \times \mathcal{U} \rightarrow \mathcal{R} \times \mathcal{C}
\label{eq:problem_formulation}
\end{equation}
where $\mathcal{C}$ represents the  classification decision and $\mathcal{R}$ represents the LLM's rationale for the decision. 

This function is implemented  through a text-to-text transformation, where prompts are crafted for a given $(p,s,u)$ using LLMs. We design the prompts to have placeholders for input data and PC-SA specific instructions, therefore reducing the auto-prompting task to optimizing the \emph{instructions} to the quality assessment LLM. To reduce the search space and parameterize the problem, we structure the instructions at PC-SA level, considering instructions to be dependent only on the SA $s_i$ and PC $p$ given by product $a$, and model it as independent from the UA value $u$. Note that UA values- titles, product descriptions, and images - are mainly what the correctness of SA $s$ gets assessed against, hence the classifier $f$ is a function of $u$, however the \emph{structure} of the prompt is modeled independently.

Thus, given the PC and SA of interest of the given product, instruction creation is  modeled  by  $\pi: \mathcal{P} \times \mathcal{S}  \rightarrow \mathcal{I}$ where  $ \mathcal{P}$ and $\mathcal{S}$  are the set of all PCs and SAs respectively, and  $\mathcal{I}$ represents the  instruction space.  These instructions are then fed into the text-to-text transformation function 
$T: \mathcal{I} \times \mathcal{U} \rightarrow \mathcal{Y}$,  to create an element of the output token space  $\mathcal{Y}$. 
The end-to-end pipeline can now be expressed as
\begin{equation*}
f(p,s,u) = g(T(\pi(p,s),u)); \quad  g: \mathcal{Y} \rightarrow \{\mathcal{R}, \mathcal{C}\}.
\end{equation*}
%
 %where  $g$  maps the generated tokens to classification labels and rationale.

% Unlike other architectures that typically require training a task-specific layer (e.g., classification) from scratch beyond the backbone model, the aforementioned formulation allows us to leverage an LLM's capacity for generating output tokens based on its pretrained knowledge, saving time and human resources. By encapsulating the PC-SA level nuances within the input instruction, the LLM can perform accurate classifications tailored to the unique characteristics of each PC-SA combination.

\subsection{CoT prompting approach}\label{sec:cot_prompting}

We adapt a Chain-of-Thought (CoT) prompting approach \cite{wei2022chain, ma2023chain}, assuming that an LLM's inherent contextual understanding from pre-training can resolve ambiguity in our task. This simplifies our general prompt function from Equation \ref{eq:problem_formulation} to $f(p, s, u) = g(T(p, s, u))$. Here, we rely on the model's implicit knowledge of the PC-SA pair $(p,s)$ to produce a high-quality transformation $T(p,s,u)$ without explicit instructions. Our zero-shot prompt incorporates intermediate reasoning steps to guide the classification, which further helps the enhanced subsequent techniques detailed next.

\subsection{Human engineered PC-SA instructions}
\label{sec:human_engineered_instructions}

% To create golden human crafted instructions for the LLM that capture nuanced requirements and store-specific expectations for each PC and the SA, we  manually create instructions for a large subset of 2,388 PC-SAs. The selection of these 2,388 PC-SAs is weighted based on frequency distribution of the product category. Formally, we use human crafted instructions for obtaining instructions $\pi (p, s)$ defined in Section \ref{sec:problem_formulation}. 

% To enhance the LLM's comprehension of domain-specific validation criteria and contextual nuances, the instruction generation framework follows a rigorous protocol involving domain experts. The process leverages quantitative attribute profiles and qualitative domain expertise to formulate detailed instructions. To optimize the human effort for instruction creation while maintaining quality, we implement a hierarchical approach, creating instructions at the attribute level rather than an individual PC-SA level.

To capture nuanced requirements and store-specific expectations at a PC-SA level, we manually engineer a set of "golden" instructions with the help of domain experts. These instructions cover 2,388 PC-SAs, which we select based on PC frequency, and serve as our implementation of the mapping function $\pi (p, s)$ defined in \S \ref{sec:problem_formulation}. To manage the significant manual effort, we adopt a hierarchical approach, authoring instructions at the more general SA level rather than for each PC-SA pair.

Each instruction provides comprehensive guidance to the LLM by including three key components: (1) a formal definition of the SA's semantic scope within e-commerce; (2) clear guidelines for handling edge cases and ambiguity; and (3) domain-specific constraints that may differ from the LLM's general pre-trained knowledge.
% \begin{itemize}
%     \item \textbf{Semantic Definition:} A formal specification of the SA's semantic scope and intended application within the e-commerce space.
%     \item \textbf{Boundary Condition Protocols:} Systematic guidelines for handling edge cases and ambiguous scenarios to help establish clear decision boundaries.
%     \item \textbf{Amazon Guidelines:} Domain-specific constraints and conventions that may diverge from general knowledge representations in pre-trained LLMs.
% \end{itemize}

% This structured approach enables consistent knowledge transfer to the LLM while maintaining scalability across diverse product categories.

\begin{algorithm}[t]
\caption{Cascaded few shot prompt generation}
\begin{algorithmic}[1]
\footnotesize
\State $\mathcal{I_\text{H}} \gets$ set of hand crafted instructions
\For{$s \in \mathcal{S}$} \Comment {iteration 1}
    \State $  \mathcal{I}_{A^{(1)}} \gets \emptyset$  \Comment {Initialize $\mathcal{I}_{A^{(1)}}$ as the set of auto-generated instructions for attribute $s$}
    \State Randomly sample $M$ product categories from $\mathcal{P}$ to get $\mathcal{P'} = \{p_1,...,p_M\}$
    \For{$p_x \in \mathcal{P'}$} \Comment{Use LLMs to generate automated instructions for $(p_x,s)$}
        \State $\mathcal{I}_{A^{(1)}}  \gets \mathcal{I}_{A^{(1)}}  \cup \text{LLM}(d(p_x), d(s), \mathcal{I}^M_\text{H})$  
    \EndFor
\EndFor 
\For{$\tau \in \{2,...,T\}$} \Comment {iteration $\tau$}
\State $\mathcal{I}_{A^{(\tau)}} \gets \emptyset$ \Comment {Initialize the  set of final PC-SA instructions}
\For{$s \in \mathcal{S}$} \quad \Comment{Use $\mathcal{I}_{A^{(\tau-1)}}$ as few shot examples to generate instructions for all $(p,s)$ pairs}
    \For{$p \in \mathcal{P}$} 
        \State $\mathcal{I}_{A^{(\tau)}} \gets  \mathcal{I}_{A^{(\tau)}}  \cup \text{LLM}(d(p), d(s), \mathcal{I}^M_{A^{(\tau-1)}})$ 
    \EndFor
\EndFor
\EndFor
\State \Return $\mathcal{I}_{A^{(T)}}$ 
\end{algorithmic}
\label{alg:prompt_generation}
\end{algorithm}

\subsection{ Cascaded PC-SA instruction generation}
\label{sec:auto_prompt_cascade}

Manually creating instructions for each PC-SA pair improves LLM comprehension but is unscalable for large catalogs that contain tens of thousands of such pairs. To address this challenge, we introduce an auto-prompt cascade for instruction generation by leveraging a minimal set of manually crafted instructions to efficiently generate nuanced, PC-SA specific instructions at scale.

Let us formally define the instruction generation problem. Let $d$ be the definition function, such that $d_{\mathcal{P}}(p): \mathcal{P} \rightarrow \mathcal{D}_\mathcal{P}$ denotes the definition of the PC $p$, and $d_{\mathcal{S}}(s): \mathcal{S} \rightarrow \mathcal{D}_\mathcal{S}$  denotes the definition of the SA $s$. We will drop $\mathcal{S}$ and $\mathcal{P}$ subscripts to simplify notation.  Let $\mathcal{I}_{\text{H}} \subset \mathcal{I}$ denote the space of human-crafted PC-SA instructions and $\mathcal{I}_{\text{A}} \subset \mathcal{I}$ denote the space of automatically generated PC-SA instructions, where $\mathcal{I}$ is the space of all instructions. 

In order to generate an instruction for a given PC-SA $(p_a,s_b)$ and capture the nuances for SA $s_b$ specific to product type $p_a$,  we propose to generate an automated PC-SA instruction $i^{'}_{p_a s_b}$ by utilizing $K$ few-shot examples over $K$ different product categories for the same SA $s_b$, denoted by ${i_{p_ks_b}} \in \mathcal{I}_{\text{H}}, 1 \leq k \leq K  $. 
This is tantamount to  modeling the instruction generation function as

% \begin{equation}
% \label{eq:ideal_few_shot}
% \begin{split}
% \pi'(d(p_a), d(s_b), i_{p_1s_b}, \ldots, i_{p_Ks_b}) &= i'_{p_as_b}, \\
% \forall a \in \{1,\cdots, |\mathcal{P}|\}, b \in \{1, \cdots, |\mathcal{S}|.\} 
% \end{split}
% \end{equation}

\begin{align}
\pi'(d(p_a), d(s_b), i_{p_1s_b}, \cdots, i_{p_Ks_b}) =i'_{p_as_b}, \notag \\
 \forall a \in \{1,\cdots,|\mathcal{P}|\}, b \in \{1,\cdots, |\mathcal{S}|\}    
\end{align}
This mapping $\pi':  \mathcal{D}_\mathcal{P} \times  \mathcal{D}_\mathcal{S} \times  \mathcal{I}_{\text{H}}^K \rightarrow \mathcal{I}_\text{A}$  
 is our proposed efficient proxy of the more ambiguous instruction mapping function  $\pi: \mathcal{P} \times \mathcal{S} \rightarrow \mathcal{I}$. However, this approach is intractable in practice as it necessitates $K$  instructions for each SA, leading to a total of $\mathcal{O}(K \times |\mathcal{S}|)$ hand-crafted instructions. 

To address this limitation, we propose a cascaded instruction generation framework where we utilize an LLM to assist the above using a much smaller set of $M$ hand-crafted instructions.  Fig~\ref{fig:two_iteration_PA_instruction} illustrates an example through the first two iterations of the cascade.
In iteration 1, we leverage $M$ manually created PC-SA instructions to generate $M$ few shot examples per SA through a mapping function $h_1: \mathcal{D}_\mathcal{P} \times \mathcal{D}_\mathcal{S} \times \mathcal{I}_{\text{H}}^M  \rightarrow \mathcal{I}_\text{A}$ defined as 
\begin{align}
\label{eq:function_iteration_1}
h_1( d(p_x), d(s_b), i_{p_1s_1},\cdots,i_{p_Ms_M})= i'_{p_xs_b},  \notag \\
 x \in \{1,\cdots,M\}, b \in \{1,\cdots,|\mathcal{S}|\}
\end{align}

In iteration 2, we  utilize $M$ automatically generated instructions as few shot examples at an SA level for the SA $s_b$ to produce PC-SA instructions for target pair $(p_a,s_b)$ through mapping function $h_2:\mathcal{D}_\mathcal{P} \times \mathcal{D}_\mathcal{S} \times \mathcal{I^\text{M}_{\text{A}}} 
 \rightarrow \mathcal{I}_\text{A}$
defined as

\begin{align}
\label{eq:function_iteration_2}
h_2( d(p_a), d(s_b), i'_{p_1s_b},\cdots,i'_{p_Ms_b}) =i'_{p_as_b},  \notag\\
a \in \{1,\cdots,|\mathcal{P}|\}, b \in \{1,\cdots,|\mathcal{S}|\}.
\end{align}

One could generally repeat this process a few rounds to iteratively refine the final instructions.
That is, utilizing the automated-instructions generated at step $\tau$ denoted by $i'^{(\tau)}_{p_1s_b}\in \mathcal{I}_{A^{(\tau)}}$, the iteration $(\tau+1)$ generations can be formalized as 
\begin{equation*}
% \label{eq:function_iteration_tau}
 h_{\tau+1}( d(p_a), d(s_b), i'^{(\tau)}_{p_1s_b},\cdots,i'^{(\tau)}_{p_Ms_b}) =i'^{(\tau+1)}_{p_as_b}  \notag \\
% & \hspace{1.3cm}
% a \in \{1,\cdots,|\mathcal{P}|\}, b \in \{1,\cdots,|\mathcal{S}|\}.
\end{equation*}
Final iteration $\tau=T$  then yields the automated instructions $i'^{(T)}_{p_1s_b}\in \mathcal{I}_{A^{(T)}}$ that will be utilized in the down-stream quality classification task. 

% {\Comment{\color{red}{{FS.} We need to change the figure to iterative . However for our experimental setup we have limited the number of iterations to 2 as detailed above and shown in Figure \ref{fig:two_iteration_PA_instruction}. }}}
This cascaded framework enables us to maintain instruction quality through hierarchical knowledge transfer and reduces manual effort from $O(K \times |\mathcal{S}|)$ to $O(M)$ by structured knowledge transfer between iterations. The pseudocode is provided in Alg. \ref{alg:prompt_generation}.

\begin{figure}[t!]
    \centering
    \includegraphics[width=0.95\columnwidth]{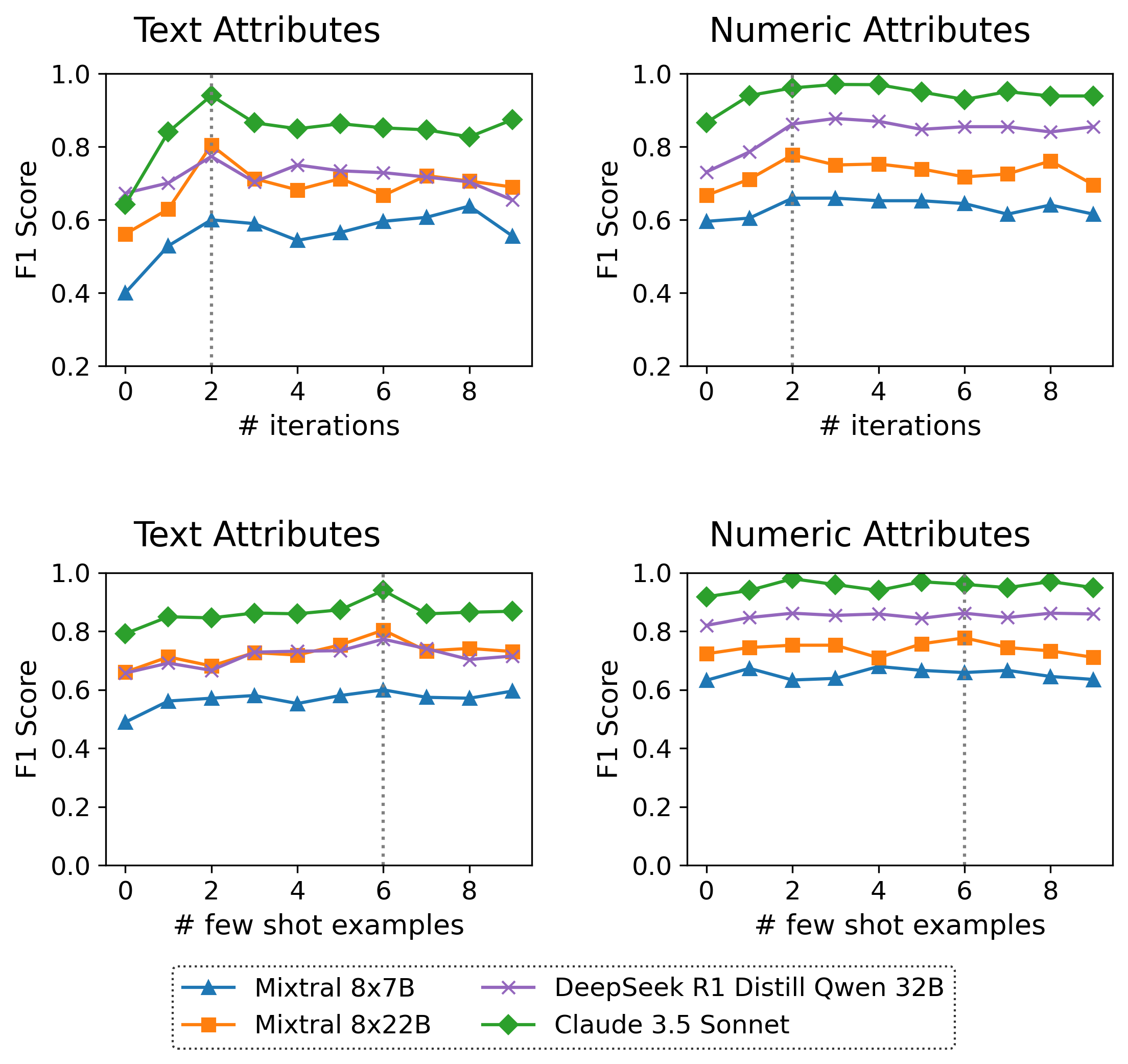}    
    % \caption{F1 score comparison on the incorrect class of three LLMs on the correctness task using Mixtral 8x7B, Mixtral 8x22B, DeepSeek R1 Distill Qwen 32B, and Claude 3.5 Sonnet. The top and bottom rows show F1 scores across the number of iterations and few shot examples respectively. Iteration 0 denotes performance of the CoT prompting. The dotted line represents best iteration across at least three of the four LLMs.}
    \caption{F1 score comparison on the incorrect class of three LLMs on the correctness task using Mixtral 8x7B, Mixtral 8x22B, DeepSeek R1 Distill Qwen 32B, and Claude 3.5 Sonnet. The top and bottom rows show F1 scores across the number of iterations and few shot examples respectively. Iteration 0 denotes performance of the CoT prompting. The dotted line represents best iteration across at least three of the four LLMs.}
    \label{fig:f1_scores_comparison}
\end{figure}

\begin{table}[t]
\centering
\scriptsize
\setlength{\tabcolsep}{1pt}
\begin{tabular}{@{}cccccc@{}}
\toprule
\textbf{Model} & \textbf{Method} & \textbf{Precision} & \textbf{Recall} & \textbf{F1 score} & \textbf{Effort} \\
\midrule
\multirow{4}{*}{Mixtral 8x7B} & Baseline & 50.08\% & 38.45\% & 43.50\% & 0 \\
  & CoT prompting & 57.67\% & 44.98\% & 50.54\% & 10 \\
  & Human engineered & 68.57\% & 51.24\% & 58.65\% & 308 \\
  & Auto prompt cascade & \textbf{70.13\%} & \textbf{52.80\%} & \textbf{60.24\%} & \textbf{3} \\

\midrule
\multirow{4}{*}{Mixtral 8x22B} & Baseline & 60.27\% & 42.61\% & 49.92\% & 0 \\
 & CoT prompting & 69.92\% & 61.32\% & 65.34\% & 10 \\
 & Human engineered & 73.05\% & 66.01\% & 68.35\% & 308 \\
 & Auto prompt cascade & \textbf{81.46\%} & \textbf{69.46\%} & \textbf{74.98\%} & \textbf{3} \\
\midrule

% \multirow{4}{*}{\shortstack{DeepSeek R1\\Distill \\ Qwen 32B}} & Baseline & 51.25\% & 75.33\% & 61.00\% & 0 \\
%  & CoT prompting & 57.50\% & 85.71\% & 68.83\% & 10 \\
%  & Human engineered & 66.10\% & 86.87\% & 75.09\% & 308 \\
%  & Auto prompt cascade & \textbf{66.30\%} & \textbf{88.21\%} & \textbf{75.70\%} & \textbf{3} \\

\multirow{4}{*}{{\shortstack{DeepSeek R1\\Distill \\ Qwen 32B}}} 
  & Baseline           & 51.25\% & 75.33\% & 61.00\% & 0 \\
  & CoT prompting      & 57.50\% & 85.71\% & 68.83\% & 10 \\
  & Human engineered   & 66.10\% & 86.87\% & 75.09\% & 308 \\
  & Auto prompt cascade & \textbf{66.30\%} & \textbf{88.21\%} & \textbf{75.70\%} & \textbf{3} \\

\midrule

\multirow{4}{*}{\shortstack{Claude 3.5 \\ Sonnet}} & Baseline & 66.00\% & 68.80\% & 67.37\% & 0 \\
 & CoT prompting & 83.95\% & 74.64\% & 79.02\% & 10 \\
 & Human engineered & 86.96\% & 80.34\% & 83.52\% & 308 \\
 & Auto prompt cascade & \textbf{92.04\%} & \textbf{90.23\%} & \textbf{91.13\%} & \textbf{3} \\

\bottomrule
\end{tabular}
\caption{Performance comparison of different prompt generation methods and models for the correctness task on english. The full table is given in Table \ref{tab:correctness_results_full}.}
\label{tab:correctness_results}
\end{table}

\section{Experimental setup}
\subsection{Tasks and datasets}
\label{sec:datasets}
% Product quality assessment can be broadly divided into two tasks: correctness and applicability. We sampled the datasets described below from an e-commerce catalog, which we used to evaluate our method. \footnote{To the best of our knowledge, there are no publicly available datasets for these tasks.}
We evaluate our method on two product quality assessment tasks, correctness and applicability, using proprietary datasets sampled from an e-commerce catalog, which we describe below. The datasets were labeled by subject matter experts and they did a 10\% random blind auditto ensure an accuracy of at least 98\% of the evaluation set. \footnote{To the best of our knowledge, there are no publicly available datasets for these tasks.}
\begin{table}[t]
\centering
\small
\setlength{\tabcolsep}{3pt}
\resizebox{\columnwidth}{!}{
\begin{tabular}{@{}lccrrrr@{}}
\toprule
\textbf{Language} & \textbf{Model} & \textbf{Prompting} & \multicolumn{3}{c}{\textbf{Incorrectness}} \\
\cmidrule(lr){4-6}
& & \textbf{technique} & \textbf{Precision} & \textbf{Recall} & \textbf{F1 score} \\
\midrule
\multirow{6}{*}{Spanish} & \multirow{3}{*}{Mixtral 8x7B} & Baseline & 34.69\% & 8.95\% & 14.23\% \\
& & CoT prompting & 56.92\% & 18.37\% & 27.78\% \\
& & Auto prompt cascade & \textbf{67.31\%} & \textbf{19.27\%} & \textbf{29.96\%} \\
\cmidrule(lr){2-6}
& \multirow{3}{*}{Mixtral 8x22B} & Baseline & 41.18\% & 10.94\% & 17.29\% \\
& & CoT prompting & \textbf{62.75\%} & 16.67\% & 26.34\% \\
& & Auto prompt cascade & 60.87\% & \textbf{21.88\%} & \textbf{32.19\%} \\
\midrule
\multirow{6}{*}{German} & \multirow{3}{*}{Mixtral 8x7B} & Baseline & 39.06\% & 12.76\% & 19.24\% \\
& & CoT prompting & 55.00\% & 22.22\% & 31.65\% \\
& & Auto prompt cascade & \textbf{61.97\%} & \textbf{22.22\%} & \textbf{32.71\%} \\
\cmidrule(lr){2-6}
& \multirow{3}{*}{Mixtral 8x22B} & Baseline & 50.98\% & 13.07\% & 20.81\% \\
& & CoT prompting & \textbf{66.20\%} & 23.62\% & 34.82\% \\
& & Auto prompt cascade & 60.64\% & \textbf{28.64\%} & \textbf{38.91\%} \\
\midrule
\multirow{6}{*}{Italian} & \multirow{3}{*}{Mixtral 8x7B} & Baseline & 20.31\% & 7.93\% & 11.41\% \\
& & CoT prompting & 62.20\% & 26.95\% & 37.61\% \\
& & Auto prompt cascade & \textbf{69.23\%} & \textbf{30.36\%} & \textbf{42.21\%} \\
\cmidrule(lr){2-6}
& \multirow{3}{*}{Mixtral 8x22B} & Baseline & 44.62\% & 17.26\% & 24.89\% \\
& & CoT prompting & 68.92\% & 30.36\% & 42.15\% \\
& & Auto prompt cascade & \textbf{78.46\%} & \textbf{30.36\%} & \textbf{43.78\%} \\
\midrule
\multirow{6}{*}{French} & \multirow{3}{*}{Mixtral 8x7B} & Baseline & 28.00\% & 7.53\% & 11.87\% \\
& & CoT prompting & 62.92\% & \textbf{29.68\%} & \textbf{40.33\%} \\
& & Auto prompt cascade & \textbf{78.12\%} & 26.32\% & 39.37\% \\
\cmidrule(lr){2-6}
& \multirow{3}{*}{Mixtral 8x22B} & Baseline & 43.86\% & 13.16\% & 20.25\% \\
& & CoT prompting & \textbf{75.00\%} & 25.13\% & 37.65\% \\
& & Auto prompt cascade & 68.97\% & \textbf{31.41\%} & \textbf{43.16\%} \\
\bottomrule
\end{tabular}
}
\caption{Performance comparison of incorrectness metrics for different prompt generation methods on multilingual datasets. The full table is given in Table \ref{tab:multilingual_results_full}.}
\label{tab:multilingual_results_compact}
\end{table}

\textbf{Correctness: } In this task, we are interested in identifying inconsistencies between UAs and SAs of a product. If an SA value is inconsistent with UAs, we call it \textit{incorrect}, otherwise it is \textit{correct}. For example, if a product description states that a jacket "keeps you dry in heavy downpours and harsh weather", but the SA "water resistance level" lists it as "water-resistant", the output would be \textit{incorrect} since water-resistant materials only protect against light rain while waterproof materials fully block moisture. 

We create a comprehensive human labeled English language dataset, covering 12,046 unique PC-SA combinations across 1,322 PCs and 1,410 SAs with both correct and incorrect labels. The dataset consists of 13,725 labels (12,148 correct, 1,577 incorrect). This is the primary task we are interested in, but we also perform experiments on other tasks that we describe below.

\textbf{Multilingual correctness: }To evaluate our framework's multilingual capabilities, we curate evaluation datasets across four major European languages. These datasets contain 1,014 labeled examples for Spanish (822 correct/192 incorrect), 1,102 for German (903 correct/199 incorrect), 1,011 for Italian (843 correct/168 incorrect), and 1,014 for French (823 correct/191 incorrect).

\textbf{Applicability: }  In this task, we are interested in identifying the SAs that are not relevant (inapplicable) for a given product. For instance, in shoes PC, the SA "heel length" is inapplicable to a pair of sneakers  while it is applicable to stilettos. Similar to the correctness-task, this is framed as a binary (Applicable/Inapplicable) classification task. Our English language dataset is comprised of 1,581 SAs where ~70\% of the instances belong to the Applicable class. We again utilize our auto prompt cascade's instructions on correctness task for applicability to demonstrate generalization across tasks.

\begin{table}[t]
\centering
\scriptsize
\setlength{\tabcolsep}{1pt}
\begin{tabular}{@{}cccccc@{}}
\toprule
\textbf{Model} & \textbf{Method} & \textbf{Precision} & \textbf{Recall} & \textbf{F1 score} & \textbf{Effort} \\
\midrule
\multirow{3}{*}{Mixtral 8x7B} & Baseline & 52\% & 50.93\% & 51.45\% & 0 \\
 & CoT prompting & 49.27\% & 55.46\% & 52.16\% & 10 \\
 & Auto prompt cascade & \textbf{54.83\%} & \textbf{58.56\%} & \textbf{56.63\%} & \textbf{3} \\
\midrule
\multirow{3}{*}{Mixtral 8x22B} & Baseline & 62.25\% & 44.54\% & 51.92\% & 0 \\
 & CoT prompting & 66.98\% & 43.92\% & 53.12\% & 10 \\
 & Auto prompt cascade & \textbf{69.30\%} & \textbf{45.15\%} & \textbf{54.62\%} & \textbf{3} \\
\bottomrule
\end{tabular}
\caption{Performance comparison of different prompt generation methods/models for the applicability task. The full table is given in Table \ref{tab:applicability_results_full}.}
\label{tab:applicability_results}
\end{table}

\subsection{Models and hyperparameters}
\label{sec:models_hyperparameters}
We employ Claude 3.5 Sonnet \cite{claude2024} to generate instructions for our auto prompt cascade and evaluate their effectiveness on four foundational LLMs: Mixtral 8x7B \cite{jiang2024mixtral}, Mixtral 8x22B \cite{mixtral2024}, DeepSeek R1 Distill Qwen 32B \cite{guo2025deepseek}, and Claude 3.5 Sonnet.

To tune the hyperparameters - \emph{iterations} $T$ and \emph{few-shot examples} $M$ - for the auto prompt cascade, we construct two balanced datasets comprising text and numeric SAs for the English correctness task, each with 250 correct and 50 incorrect samples. We first determine the optimal value of T by conducting experiments for $T=0\ldots6$, where $T=0$ is a baseline without auto prompt cascade instructions, i.e. the CoT prompt. Using this optimal $T$, we then tune for $M$. To ensure robustness and generalization, we select the final hyperparameter values for $T$ and $M$ that yield optimal performance across a majority of tested LLMs.

\subsection{Metrics and evaluation}
\label{sec:metrics_evaluation}
For our binary classification tasks (Correct/Incorrect or Applicable/Inapplicable), we report precision, recall, and F1 score. We prioritize metrics on the negative class for hyperparameter tuning, reflecting our task's focus. To assess scalability, we measure \textit{Effort}, the time in minutes required for manual prompt engineering per PC-SA.

We evaluate four methods: (1) a baseline using vanilla prompting, (2) CoT prompting (\S \ref{sec:cot_prompting}), (3) human-engineered instructions (\S \ref{sec:human_engineered_instructions}), and (4) our auto-prompt cascade (\S \ref{sec:auto_prompt_cascade}). Due to the time-intensive nature of method (3), we apply it only to our primary task of English correctness. Similarly, to manage computational cost for the multilingual and applicability tasks, we report results on two LLMs: Mixtral 8x7B and Mixtral 8x22B. We show the prompt templates used in \S \ref{sec:prompt_templates}.

\section{Results}
\label{sec:results}

\begin{table*}[ht]
\centering
\scriptsize
\begin{tabular}{|p{0.05\textwidth}|p{0.1\textwidth}|p{0.4\textwidth}|p{0.35\textwidth}|}
\hline
\textbf{PC} & \textbf{SA (value)} & \textbf{CoT (abridged reasoning)} & \textbf{APC (abridged reasoning)} \\
\hline
Walking stick & base material (rubber) & Incorrect: The product data mentions that the walking stick is made of 'Huangtang wood' and has a 'steel spike' and 'metal-reinforced rubber tip cover'. There is no mention of the walking stick being made of rubber material. & Correct: The product data mentions a 'metal-reinforced removable rubber tip cover' for the walking stick, which implies that the base material that comes into contact with the ground is rubber. \\
\hline
\end{tabular}
\caption{Qualitative example contrasting CoT and auto-prompt cascade predictions. The table presents abridged reasoning segments that directly contributed to the correct decision, with complete rationales and further examples provided in \S\ref{sec:qualitative_results}.}
\label{tab:compact_qualitative}
\end{table*}

\textbf{Determining hyperparamaters: } We determine the optimal number of iterations, $T$, using our hyperparameter tuning datasets (\S \ref{sec:models_hyperparameters}) for the correctness task. As plotted in Fig \ref{fig:f1_scores_comparison}, F1 scores peak at $T$=2. The performance gain from $T$=1 to $T$=2 shows the model's ability to capture nuances through iterative refinement. For instance, the instruction for a walking stick's base material evolves from a generic definition at $T$=1: (\textit{A base material for a walking stick refers to the primary material used to construct the base of the walking stick, which is typically wood, aluminum, carbon fiber, or other sturdy materials})
to a more contextualized instruction at $T$=2: (\textit{Base material refers to the material that makes up the bottom part of a walking stick, which comes into contact with the ground and provides stability and traction when using the stick}). This improvement stems from the model first integrating domain-specific metadata at $T$=1, then performing SA-level refinement at $T$=2 for finer distinctions.

Our method is also robust to the number of few-shot examples, $M$. Fig \ref{fig:f1_scores_comparison} shows that with the optimal $T$=2, performance is steady across various values of $M$, with most models showing peak F1 score at $M$=6. Given these observations, we select $T$=2 and $M$=6 for our auto-prompt cascade in all subsequent experiments.

% The observed performance plateau beyond iteration two, coupled with consistently higher accuracy on numeric attributes compared to textual attributes across all evaluated models, indicates optimal instruction synthesis at iteration two. This empirical evidence supports our implementation of a two-iteration system architecture. The superior performance on numeric validation tasks suggests that structured numerical constraints are more effectively processed by LLMs compared to semantic text validation, potentially due to the inherent precision and well-defined boundaries of numerical attributes.

\textbf{Correctness: } As shown in Table \ref{tab:correctness_results}, the auto prompt cascade consistently improves F1 scores for incorrect SA detection across all models. With Claude 3.5 Sonnet, it achieves a 91.13\% F1 score, a 23.76\% improvement over the baseline. We observe similar gains with DeepSeek R1 Distill Qwen 32B (61.00\% to 75.70\%), Mixtral 8x22B (from 49.92\% to 74.98\%), and Mixtral 8x7B (from 43.50\% to 60.24\%).

Our approach also significantly reduces human effort. The human-engineered method requires 308 minutes per PC-SA to cover only 2,388 PC-SAs, whereas our auto prompt cascade handles all 12,046 PC-SAs in just 3 minutes per PC-SA. This represents a 99\% reduction in human time for superior performance. In comparison, the CoT approach requires 10 minutes per PC-SA but yields consistently lower performance with no PC-SA specific instructions.

\textbf{Multilingual correctness: } We use Mixtral 8x7B and Mixtral 8x22B to compare our auto prompt cascade with CoT and vanilla prompting approaches. Table \ref{tab:multilingual_results_compact} suggests that our approach improves the F1 score across all four languages - Spanish, German, Italian, and French by 10-30\% over the baseline, demonstrating the cross-lingual generalization capability of our approach.

{\bf Applicability:} Table \ref{tab:applicability_results} shows consistent improvements across F1 score for the Inapplicable class, from  51.45\% to 56.63\% for Mixtral 8x7B, and from 51.92\% to 54.62\% for Mixtral 8x22B, demonstrating the generalization capability of the automatically-generated instructions of the auto prompt cascade approach across correctness and applicability tasks.  

\textbf{Qualitative examples:}  To illustrate the effectiveness of our approach, we present a qualitative example where auto prompt cascade (APC) instructions helped the model correct its initial assessment error for the correctness task in Table \ref{tab:compact_qualitative}. The table provides an abridged version of the reasoning that led to this correction, while full reasoning details and additional examples are discussed in \S\ref{sec:qualitative_results}. The PC is a walking stick and the value of the base material is rubber, which is correct in ground truth. The CoT prompt implicitly interprets base material as the material of the entire stick. Seeing “Huangtang wood” and a “steel spike,” it concludes the test value “rubber” contradicts the product data and predicts Incorrect. 

The APC instruction generated is \textit{Base material refers to the material that makes up the bottom part of a walking stick, which comes into contact with the ground and provides stability and traction when using the stick.} With this disambiguation, the model’s evidence retrieval shifts to the phrase “metal-reinforced removable rubber tip cover,” and the rationale updates accordingly, yielding Correct. This example illustrates how the learned instruction resolves an attribute ambiguity at a PC level.

\textbf{Cost: } Our method generates instructions for each PC-SA combination offline and the generated instruction for the PC-SA is then used for all the items in the PC. Hence, the number of LLM inferences required for our method is linearly proportional to the number of PC-SAs (tens of thousands) and not the number of items in the catalog (hundreds of millions). 

The final results in the paper needed only 20,506 instruction generation LLM calls. We generated the instructions using Claude 3.5 Sonnet, with an average of 2717 input tokens and 113 output tokens per LLM call. In comparison to CoT, our approach consumes the extra computational cost of \$ 0.0097/PC-SA for creating the instructions, which is negligible. The inference costs on the downstream quality classification tasks for CoT and our approach are comparable as both use the same underlying LLM.

\textbf{Statistical signifcance: } We performed a statistical evaluation of the results, showing that the performance gains are statistically significant in the vast majority of cases. We assessed the significance using a paired subsampled bootstrap test with 5,000 class-balanced 80\% draws without replacement, applying a plus-one correction. The detailed $\Delta$ F1 scores and the p values are shown in \S\ref{sec:statistical_significance}.

\section{Conclusion and future work}
\label{sec:conclusion}
%Looking ahead, several promising research directions could further enhance the capabilities and efficiency of our framework, including  developing quantitative metrics to explicitly evaluate instruction quality in terms of clarity, consistency, and coverage of edge cases, as well as  development of dynamic instruction update mechanisms that can automatically refine and improve instructions based on model performance feedback, enhancing efficiency and effectiveness of the framework. 
% Our work also highlights the need for new quantitative metrics to systematically evaluate instructions themselves, beyond downstream task performance. Such metrics would assess clarity, consistency across scenarios, and coverage of unusual or edge cases.
We propose an auto-prompt cascade for assessing product quality with LLMs - the first scalable approach for multilingual catalog quality assessment. It expands a small set of human-crafted instructions (6) to over 12,000 PC-SA-specific prompts, reducing manual effort by 99\%. Our cascade demonstrates improvement of classification performance across two tasks and generalizes to five languages and multiple LLMs. While applied to product quality, it can be extended to product categorization and attribute generation. This work also underscores the need for quantitative metrics to evaluate instruction quality directly, measuring clarity, consistency, and handling of edge cases.

\section{Limitations}
\label{sec:limitations}

While our auto-prompt cascade demonstrates significant improvements, we acknowledge some limitations that offer avenues for future research. 

Firstly, the evaluation of the generated instructions is currently extrinsic, measured only by their impact on downstream quality tasks. As noted in our conclusion, there could be a need for intrinsic, quantitative metrics to directly assess the quality of the instructions themselves - for instance, by measuring their clarity, consistency, and ability to handle edge cases without relying on models run on downstream tasks. 

Secondly, while we demonstrate strong performance on e-commerce quality tasks like correctness and applicability, the framework's effectiveness on other e-commerce tasks, such as attribute value generation or product categorization, remains an open question. Extending and validating the auto prompt cascade for these different tasks is a key direction for future work.

Finally, an additional limitation pertains to the proprietary nature of our dataset. To alleviate this, we provide a thorough description of the quality assessment tasks and a detailed statistical breakdown of the composition of the datasets.

\color{black}

\bibliographystyle{acl_natbib}
\bibliography{custom}
\clearpage

\appendix

\section{Prompt templates}
\label{sec:prompt_templates}
The following subsections contain the prompt templates that we use for our experiments. 

\subsection{Baseline prompt template}
\label{sec:baseline_prompt}
We use the prompt template shown in Figure \ref{fig:baseline_prompt_template} for the baseline experiment as detailed in \S \ref{sec:metrics_evaluation}. We use the same baseline prompt for all PC-SA pairs. Given below is an example of the template for the SA - base material. This template instructs the model to adopt an auditor persona and perform a binary classification (\textit{Correct} or \textit{Incorrect}). Note that it defines \textit{Correct} as the absence of a contradiction, rather than requiring explicit confirmation, which is inline with our correctness problem defined in \S\ref{sec:datasets}.

\begin{figure*}[h!]
    \begin{tcolorbox}[colback=yellow!20, colframe=black, coltext=black, fontupper=\ttfamily]  
    
    You are an auditor for an e-commerce store. You are given a product and its data below. You will also be given a test value for 'base material'. 
    \\ \\
    <Product data goes here.>
    \\ \\
    The test value for 'base material' is 'rubber'.
    Based on the given product data, you have to say if the test value is 'Correct' or 'Incorrect'. If the product data does not contradict the given value, your prediction should be 'Correct'.
    \\ \\
    Output the results in the following output format. \\
    <Output format>
    \end{tcolorbox}
    \caption{Baseline prompt template for the attribute \textit{base material}.}
    \label{fig:baseline_prompt_template}
\end{figure*}

\subsection{CoT prompt template}
\label{sec:cot_example}

For the CoT experiment as detailed in \S \ref{sec:metrics_evaluation}, we use different prompts for different groups of SAs. The prompt template shown in Figure \ref{fig:cot_prompt_template} is an example used for the attribute \textit{base material} in the \textit{walking stick} product category. 

Unlike the baseline, this CoT prompt is highly structured with explicit sections for introduction, data, rules, test value, and output format. The key component is the \textit{Rules} section, which contains a placeholder for a sequence of reasoning steps that guide the LLM make accurate classifications. This design encourages the model to "think systematically" by following a prescribed logical flow before making its final assessment, aiming to improve accuracy on nuanced and complex cases.

\begin{figure*}[h!]
    \begin{tcolorbox}[colback=yellow!20, colframe=black, coltext=black, fontupper=\ttfamily]  
    \#\#\# Introduction: \\
    You are an auditor for an e-commerce store.  You are given a product and its data below. You will also be given a test value for 'base material'. \\
    Please classify the value as 'Correct' or 'Incorrect' based on the rules given below. \\
    
    \#\#\# Product  data: \\
    Given below is the product  data. \\
    <Product data goes here.> \\ 
      
    \#\#\# Rules: \\
    To ensure accurate predictions, adhere to the following rules in sequence and think systematically before responding: \\
    <CoT rules go here.> 
    \\ \\
    \#\#\# Test value: \\
    Now verify the test value of the attribute 'base material': 'rubber'.
    \\ \\
    \#\#\# Output format: \\
    Output the results in the following output format. \\
    <Output format>
    \end{tcolorbox}
    \caption{CoT prompt template for the product category \textit{walking stick} and attribute \textit{base material}.}
    \label{fig:cot_prompt_template}
\end{figure*}

\subsection{Auto prompt cascade prompt template}
Finally, for our proposed auto-prompt cascade method, we augment the CoT prompt with an additional \textit{Instruction} field. This section contains the nuanced, domain-specific definition that is automatically generated by our cascade. The prompt template shown in Figure \ref{fig:auto_prompt_cascade_prompt_template} illustrates this for the base material attribute of a walking stick. Notice it is identical to the CoT template in \S\ref{sec:cot_example}, with the exception of the crucial auto-generated instruction.

\begin{figure*}[h!]
    \begin{tcolorbox}[colback=yellow!20, colframe=black, coltext=black, fontupper=\ttfamily]  
    \#\#\# Introduction: \\
    You are an auditor for an e-commerce store.  You are given a product and its data below. You will also be given a test value for 'base material'. \\
    Go through the instruction to understand what 'base material' means in context of this product. Please classify the value as 'Correct' or 'Incorrect' based on the rules given below. \\ 
    
    \#\#\# Product  data: \\
    Given below is the product  data. \\
    <Product data goes here.> \\ 
      
    \#\#\# Rules: \\
    To ensure accurate predictions, adhere to the following rules in sequence and think systematically before responding: \\
    <CoT rules go here.> \\
    
    \#\#\# Test value: \\
    Now verify the test value of the attribute 'base material': 'rubber'.
    \\ 
    
    \#\#\# Instruction: \\
    Here is some additional information about 'base material' to help you make highly accurate classifications. \\
    In your reasoning, explain how you applied this information to reach your conclusion.
    \\ \\ 
    Base material refers to the material that makes up the bottom part of a walking stick, which comes into contact with the ground and provides stability and traction when using the stick.
    \\ \\ 
    \#\#\# Test value: \\
    Now verify the test value of the attribute 'base material': 'rubber'.
    \\ \\
    \#\#\# Output format: \\
    Output the results in the following output format. \\
    <Output format>
    \end{tcolorbox}
    \caption{Auto prompt cascade prompt template for the product category \textit{walking stick} and attribute \textit{base material}. We add the auto generated instruction to the walking stick-base material PC-SA pair to the CoT prompt template.}
    \label{fig:auto_prompt_cascade_prompt_template}
\end{figure*}

This final template directly addresses the core challenge of domain-specific knowledge gaps. By inserting the auto-generated instruction, we provide the LLM with a precise, contextual definition for the PC-SA pair. Furthermore, the instruction forces the model to ground its reasoning in this new information, making its decision making process more transparent and aligned with the specific requirements of the e-commerce catalog.

\color{black}
\section{Hyperparameter Analysis}
\label{sec:hyperparameter_experiments}
\begin{figure*}[t!]
    \centering
    \includegraphics[width=0.9\textwidth]{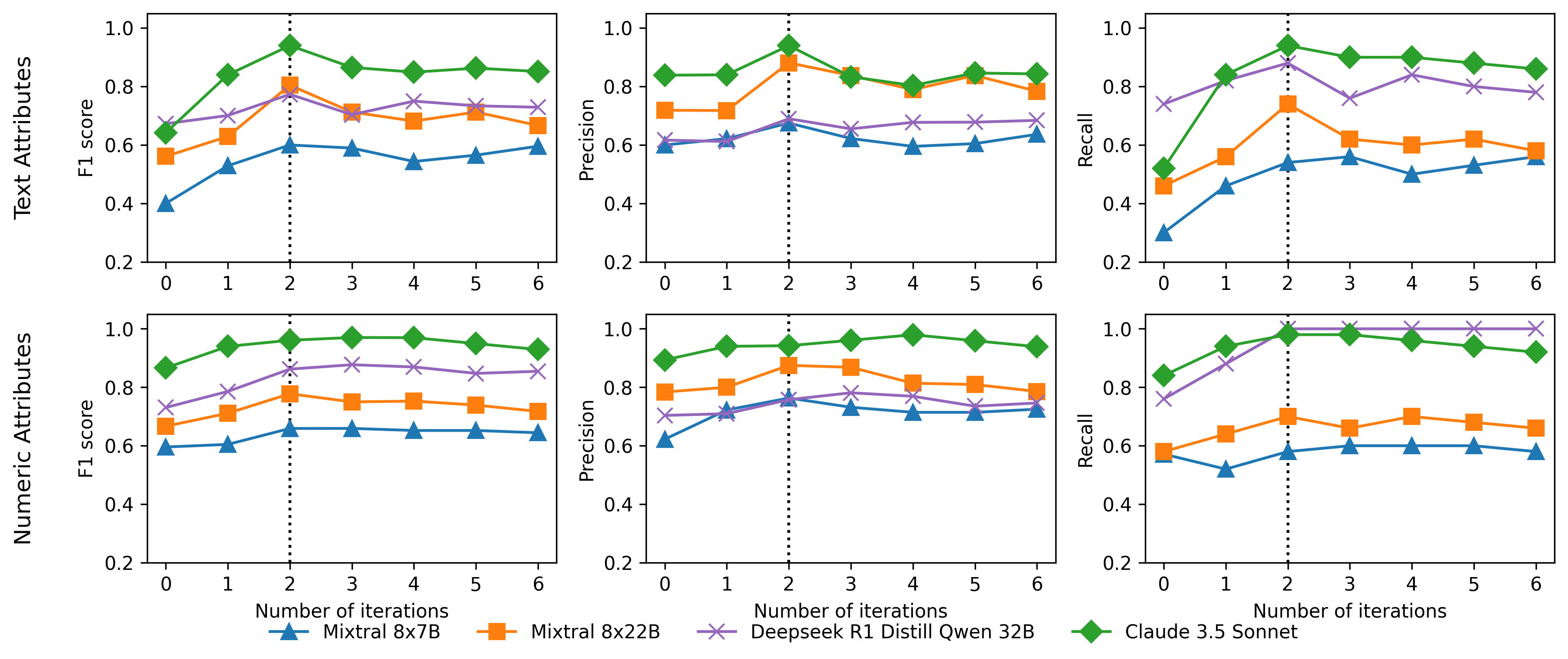}    
    % \caption{Performance comparison on the incorrect class of four LLMs across the number of iterations used for instruction generation on the correctness task using Mixtral 8x7B, Mixtral 8x22B, DeepSeek R1 Distill Qwen 32B, and Claude 3.5 Sonnet. The top and bottom rows show performance on text and numeric attributes respectively. Iteration 0 denotes performance before the auto-prompt cascade. The dotted line represents best iteration across majority of the LLMs.}
    \caption{Performance comparison on the incorrect class of four LLMs across the number of iterations used for instruction generation on the correctness task using Mixtral 8x7B, Mixtral 8x22B, DeepSeek R1 Distill Qwen 32B, and Claude 3.5 Sonnet. The top and bottom rows show performance on text and numeric attributes respectively. Iteration 0 denotes performance before the auto-prompt cascade. The dotted line represents best iteration across majority of the LLMs.}

    \label{fig:model_performance_across_iters}
\end{figure*}

This section presents a detailed analysis of the hyperparameter tuning experiments for our auto-prompt cascade method. As introduced in \S \ref{sec:models_hyperparameters}, Figure \ref{fig:model_performance_across_iters} illustrates the precision, recall, and F1 score of the incorrect class across multiple iterations, $T$, of the auto-prompt cascade. Iteration 0 represents the baseline performance obtained with standard CoT prompting, prior to the application of the auto-prompt cascade. The dotted line in each subplot denotes the empirically determined optimal iteration, defined as the earliest iteration where at least three of the four evaluated LLMs achieve their peak performance for the corresponding metric. Notably, our method consistently demonstrates optimal performance around the second iteration, indicating rapid convergence and highlighting the efficiency of the auto-prompt cascade in improving model performance.

\begin{figure*}[ht]
    \centering
    \includegraphics[width=0.9\textwidth]{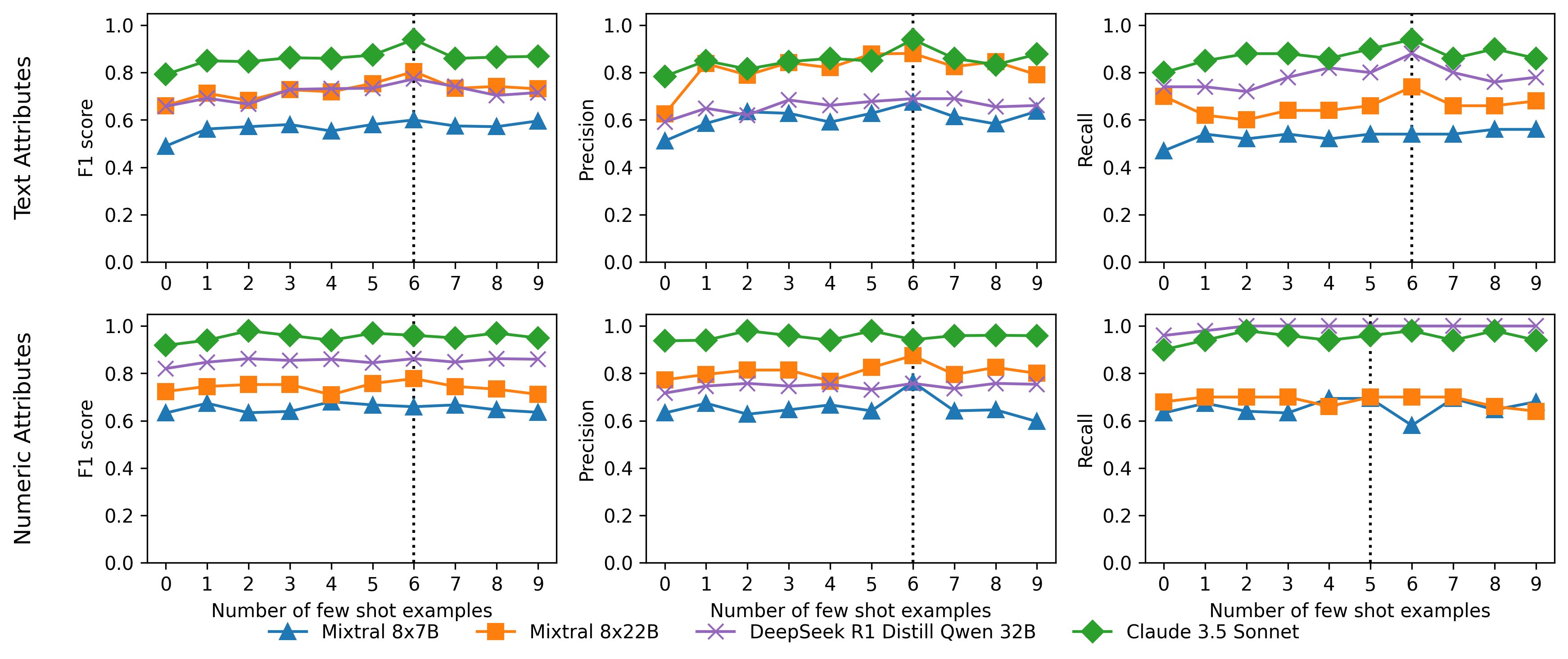}    
    % \caption{Performance comparison on the incorrect class across the number of few shot examples used for instruction generation on the correctness task, using Mixtral 8x7B, Mixtral 8x22B, DeepSeek R1 Distill Qwen 32B, and Claude 3.5 Sonnet. The top and bottom rows show performance on text and numeric attributes respectively. Iteration 0 signifies instruction generation using only the PC and SA definitions. The dotted line represents optimal selection of few-shot examples across majority of the LLMs.}
    \caption{Performance comparison on the incorrect class across the number of few shot examples used for instruction generation on the correctness task, using Mixtral 8x7B, Mixtral 8x22B, DeepSeek R1 Distill Qwen 32B, and Claude 3.5 Sonnet. The top and bottom rows show performance on text and numeric attributes respectively. Iteration 0 signifies instruction generation using only the PC and SA definitions. The dotted line represents optimal selection of few-shot examples across majority of the LLMs.}
    \label{fig:model_performance_few_shot}
\end{figure*}

After determining the optimal number of iterations, $T=2$, we proceed with hyperparameter tuning on the same datasets to ascertain the optimal number of few-shot examples, $M$. Figure \ref{fig:model_performance_few_shot} illustrates the impact of varying $M$ on the precision, recall, and F1 score for the incorrect class. Iteration 0 in this context signifies instruction generation based solely on the PC and SA definitions, without any few-shot examples. The dotted line in each subplot marks the optimal selection of few-shot examples, corresponding to the earliest point where the majority of the evaluated LLMs exhibit peak performance. Observing the results presented in Figure \ref{fig:model_performance_across_iters}, it is evident that the majority of the subplots (5 out of 6) demonstrate optimal performance for the LLMs when $M=6$ few-shot examples are utilized. This consistent trend across diverse metrics and models strongly indicates that $M=6$ represents the most effective configuration for instruction generation within our auto-prompt cascade method. Consequently, we adopted this value for $M$ in our experiments involving the auto-prompt cascade.

\section{Full results}
\label{sec:full_results}
In this section, we present the full results of the following tables - Table \ref{tab:correctness_results}, Table \ref{tab:multilingual_results_compact}, and Table \ref{tab:applicability_results}. In these tables, we show the precision/recall/F1 score metrics on the positive class as well ("correct" for the correctness task and "applicable" for the applicability task). We also provide the numbers for accuracy on the respective datasets and the numbers of unique PC-SA combinations which had instructions generated for a particular experiment.

\subsection{English correctness results}
\label{sec:en_correctness_results}
As detailed in section \ref{sec:datasets}, correctness in english is our main dataset with 13,725 labels (12,148 correct, 1,577 incorrect) across 12,046 unique PC-SA combinations. Table \ref{tab:correctness_results_full} is the full version of Table \ref{tab:correctness_results}. It presents the performance comparison of four LLMs, Mixtral 8x7B, Mixtral 8x22B, DeepSeek R1 Distill Qwen 32B, and Claude 3.5 Sonnet, across four prompting methods: baseline, Chain-of-Thought (CoT) prompting, human-engineered prompts, and our proposed auto-prompt cascade. Across all models and metrics, the auto-prompt cascade consistently outperforms other methods. For instance, on Mixtral 8x22B, it achieves the highest incorrectness F1 score of 74.98\% and correctness F1 score of 97.02\%, with an overall accuracy of 94.37\%, surpassing both CoT prompting and human-engineered prompts. Notably, the cascade scales to 12,046 unique PC-SA combinations with just 3 minutes of human effort per unique PC-SA combination, compared to 5.1 hours (308 minutes) for human-engineered prompts. This trend holds consistently across other models, with Claude 3.5 Sonnet reaching a correctness F1 of 98.86\% and an accuracy of 97.62\%. 

When comparing model performance under the same prompting method, larger models consistently outperform smaller ones. For instance, under CoT prompting, Mixtral 8x22B outperforms Mixtral 8x7B on correctness F1 score (95.81\% vs. 94.33\%), incorrectness F1 score (65.34\% vs 50.54\%), and accuracy (92.21\% vs. 89.50\%). Likewise, human-engineered prompts show progressively better results on larger models, with Claude 3.5 Sonnet achieving the highest correctness F1 of 97.95\% under this setting. Under the auto-prompt cascade, Claude 3.5 Sonnet consistently leads with the best overall performance (correctness F1 98.86\%, accuracy 97.62\%), followed by Mixtral 8x22B and DeepSeek R1 Distill Qwen 32B. Interestingly, while DeepSeek R1 Distill Qwen 32B lags behind on baseline and CoT prompting, its performance under the auto-prompt cascade improves substantially, achieving a correctness F1 of 96.24\% and accuracy of 93.17\%, indicating that prompt quality has an outsized impact on lower-capacity models. These results demonstrate that the auto-prompt cascade consistently boosts performance across model sizes and architectures, narrowing the gap between smaller and larger models while dramatically reducing manual effort.

\begin{table*}[t]
\centering
\scriptsize
\setlength{\tabcolsep}{3pt}
\begin{tabular}{@{}ccccccccccc@{}}
\toprule
\textbf{Model} & \textbf{Method} & \multicolumn{3}{c}{\textbf{Incorrectness}} & \multicolumn{3}{c}{\textbf{Correctness}} & \textbf{Accuracy} & \textbf{Unique PC-SA} & \textbf{Effort } \\
\cmidrule(lr){3-5} \cmidrule(lr){6-8}
 &  & \textbf{Precision} & \textbf{Recall} & \textbf{F1 score} & \textbf{Precision} & \textbf{Recall} & \textbf{F1 score} & & \textbf{Count} &  \\
\midrule

\multirow{4}{*}{Mixtral 8x7B} & Baseline & 50.08\% & 38.45\% & 43.50\% & 92.23\% & 95.01\% & 93.60\% & 88.18\% & 0 & 0 \\
  & CoT prompting & 57.67\% & 44.98\% & 50.54\% & 93.01\% & 95.68\% & 94.33\% & 89.50\% & 0 & 10 \\
  & Human engineered & 68.57\% & 51.24\% & 58.65\% & 93.89\% & 96.96\% & 95.40\% & 91.40\% & 2388 & 308 \\
  & Auto prompt cascade & \textbf{70.13\%} & \textbf{52.80\%} & \textbf{60.24\%} & \textbf{94.05\%} & \textbf{97.07\%} & \textbf{95.54\%} & \textbf{91.66\%} & 12046 & \textbf{3} \\

\midrule
\multirow{4}{*}{Mixtral 8x22B} & Baseline & 60.27\% & 42.61\% & 49.92\% & 92.82\% & 96.35\% & 94.55\% & 89.87\% & 0 & 0 \\
 & CoT prompting & 69.92\% & 61.32\% & 65.34\% & 95.06\% & 96.58\% & 95.81\% & 92.21\% & 0 & 10 \\
 & Human engineered & 73.05\% & 66.01\% & 68.35\% & 95.64\% & 96.84\% & 96.24\% & 92.98\% & 2388 & 308 \\
 & Auto prompt cascade & \textbf{81.46\%} & \textbf{69.46\%} &\textbf{74.98\%} & \textbf{96.11\%} & \textbf{97.95\%} & \textbf{97.02\%} & \textbf{94.37\%} & 12046 & \textbf{3} \\
\midrule

\multirow{4}{*}{\shortstack{DeepSeek R1\\Distill Qwen 32B}} & Baseline & 51.25\% & 75.33\% & 61.00\% & 96.59\% & 90.70\% & 93.55\% & 88.62\% & 0 & 0 \\
  & CoT prompting & 57.50\% & 85.71\% & 68.83\% & 93.02\% & 91.78\% & 94.80\% & 90.77\% & 0 & 10 \\
  & Human engineered & 66.10\% & 86.87\% & 75.09\% & 98.22\% & 94.21\% & 96.17\% & 93.07\% & 2388 & 308 \\
  & Auto prompt cascade & \textbf{66.30\%} & \textbf{88.21\%} & \textbf{75.70\%} & \textbf{98.40\%} & \textbf{94.17\%} & \textbf{96.24\%} & \textbf{93.17\%} & 12046 & \textbf{3} \\
\midrule

\multirow{4}{*}{Claude 3.5 Sonnet} & Baseline & 66.00\% & 68.80\% & 67.37\% & 95.93\% & 95.40\% & 95.66\% & 92.03\% & 0 & 0 \\
 & CoT prompting & 83.95\% & 74.64\% & 79.02\% & 96.75\% & 98.15\% & 97.44\% & 95.13\% & 0 & 10 \\
 & Human engineered & 86.96\% & 80.34\% & 83.52\% & 97.47\% & 98.44\% & 97.95\% & 96.04\% & 2388 & 308 \\
 & Auto prompt cascade & \textbf{92.04\%} & \textbf{90.23\%} & \textbf{91.13\%} & \textbf{98.74\%} & \textbf{98.99\%} & \textbf{98.86\%} & \textbf{97.62\%} & 12046 & \textbf{3} \\

\bottomrule
\end{tabular}
\caption{Performance comparison of different prompt generation methods and models for the correctness task on the english language.}
\label{tab:correctness_results_full}
\end{table*}

\subsection{Multilingual correctness results}
\label{sec:multilingual_correctness}

We mention in \S\ref{sec:datasets} that we also test our auto prompt cascade on four european language datasets -  Spanish, German, Italian, and French. These datasets contain 752, 799, 734, and 793 PC-SA pairs respectively. Table~\ref{tab:multilingual_results_full}, the full version of Table~\ref{tab:multilingual_results_compact}, reports the precision, recall, F1 score, and accuracy for both correctness and incorrectness classifications using Mixtral 8x7B and Mixtral 8x22B across baseline, CoT prompting, and auto-prompt cascade methods. In all cases, the auto-prompt cascade outperforms both baseline and CoT prompting on the key metrics of Incorrectness F1 score, Correctness F1 score, and Accuracy. For example, in Spanish, Mixtral 8x7B’s Incorrectness F1 score improves from 14.23\% (baseline) and 27.78\% (CoT) to 29.96\% with the auto prompt cascade, while Correctness F1 rises from 88.37\% to 90.20\%, and accuracy increases from 79.52\% to 81.37\%. 

When comparing models within each prompting technique, Mixtral 8x22B consistently outperforms Mixtral 8x7B on Incorrectness F1, Correctness F1, and Accuracy for both baseline and CoT prompting across all languages. For instance, under CoT prompting in Italian, Mixtral 8x22B achieves an Incorrectness F1 score of 42.15\%, Correctness F1 score of 92.13\%, and accuracy of 83.18\%, compared to 37.61\%, 91.44\%, and 82.32\% for 8x7B, respectively. However, the auto-prompt cascade substantially narrows this performance gap. In French, Mixtral 8x7B with the cascade achieves an Incorrectness F1 of 39.37\%, Correctness F1 score of 91.23\%, and accuracy of 81.59\%, approaching the CoT-prompted 8x22B values of 43.16\%, 90.97\%, and 81.36\%. In several cases, the cascade-augmented 8x7B even outperforms CoT-prompted 8x22B, such as in Italian where it reaches 42.21\% Incorrectness F1 score, 92.20\% Correctness F1 score, and 82.90\% accuracy. These results reaffirm the cascade’s strong generalizability and its ability to consistently improve both correctness and error detection across models and languages while reducing manual effort.

\begin{table*}[!htbp]
\centering
\scriptsize
\setlength{\tabcolsep}{3pt}
\begin{tabular}{@{}ccccccccccc@{}}
\toprule
\textbf{Model} & \textbf{Prompting} & \multicolumn{3}{c}{\textbf{Incorrectness}} & \multicolumn{3}{c}{\textbf{Correctness}} & \textbf{Accuracy} & \textbf{Unique PC-SAs} & \textbf{Effort} \\
\cmidrule(lr){3-5} \cmidrule(lr){6-8}
& \textbf{technique} & \textbf{Precision} & \textbf{Recall} & \textbf{F1 score} & \textbf{Precision} & \textbf{Recall} & \textbf{F1 score} & & \textbf{with inst.} &  \\
\midrule
\multicolumn{11}{c}{\textbf{Spanish}} \\
\midrule
\multirow{3}{*}{Mixtral 8x7B} & Baseline & 34.69\% & 8.95\% & 14.23\% & 81.83\% & 96.05\% & 88.37\% & 79.52\% & 0 & 0 \\
& CoT prompting & 56.92\% & 18.37\% & 27.78\% & 83.51\% & 96.56\% & 89.56\% & 80.40\% & 0 & 10 \\
& Auto prompt cascade & \textbf{67.31\%} & \textbf{19.27\%} & \textbf{29.96\%} & \textbf{83.61\%} & \textbf{97.91\%} & \textbf{90.20\%} & \textbf{81.37\%} & 752 & \textbf{3} \\
\midrule
\multirow{3}{*}{Mixtral 8x22B} & Baseline & 41.18\% & 10.94\% & 17.29\% & 82.11\% & 96.32\% & 88.65\% & 80.04\% & 0 & 0 \\
& CoT prompting & \textbf{62.75\%} & 16.67\% & 26.34\% & 83.28\% & \textbf{97.67\%} & 89.90\% & 80.85\% & 0 & 10 \\
& Auto prompt cascade & 60.87\% & \textbf{21.88\%} & \textbf{32.19\%} & \textbf{84.03\%} & 96.69\% & \textbf{89.92\%} & \textbf{81.05\%} & 752 & \textbf{3} \\
\midrule
\multicolumn{11}{c}{\textbf{German}} \\
\midrule
\multirow{3}{*}{Mixtral 8x7B} & Baseline & 39.06\% & 12.76\% & 19.24\% & 81.55\% & 95.09\% & 87.80\% & 78.81\% & 0 & 0 \\
& CoT prompting & 55.00\% & 22.22\% & 31.65\% & 83.39\% & 95.55\% & 89.06\% & 79.05\% & 0 & 10 \\
& Auto prompt cascade & \textbf{61.97\%} & 22.22\% & \textbf{32.71\%} & \textbf{83.49\%} & \textbf{96.65\%} & \textbf{89.59\%} & \textbf{79.88\%} & 799 & \textbf{3} \\
\midrule
\multirow{3}{*}{Mixtral 8x22B} & Baseline & 50.98\% & 13.07\% & 20.81\% & 81.96\% & \textbf{96.92\%} & 88.81\% & \textbf{80.40\%} & 0 & 0 \\
& CoT prompting & \textbf{66.20\%} & 23.62\% & 34.82\% & 83.85\% & 97.05\% & \textbf{89.97\%} & 80.53\% & 0 & 10 \\
& Auto prompt cascade & 60.64\% & \textbf{28.64\%} & \textbf{38.91\%} & \textbf{84.53\%} & 95.45\% & 89.66\% & 80.24\% & 799 & \textbf{3} \\
\midrule
\multicolumn{11}{c}{\textbf{Italian}} \\
\midrule
\multirow{3}{*}{Mixtral 8x7B} & Baseline & 20.31\% & 7.93\% & 11.41\% & 83.68\% & 93.82\% & 88.46\% & 79.58\% & 0 & 0 \\
& CoT prompting & 62.20\% & 26.95\% & 37.61\% & 87.04\% & 96.31\% & 91.44\% & 82.32\% & 0 & 10 \\
& Auto prompt cascade & \textbf{69.23\%} & \textbf{30.36\%} & \textbf{42.21\%} & \textbf{87.35\%} & \textbf{97.62\%} & \textbf{92.20\%} & \textbf{82.90\%} & 734 & \textbf{3} \\
\midrule
\multirow{3}{*}{Mixtral 8x22B} & Baseline & 44.62\% & 17.26\% & 24.89\% & 85.26\% & 95.71\% & 90.18\% & 82.64\% & 0 & 0 \\
& CoT prompting & 68.92\% & 30.36\% & 42.15\% & 87.51\% & 97.27\% & 92.13\% & 83.18\% & 0 & 10 \\
& Auto prompt cascade & \textbf{78.46\%} & 30.36\% & \textbf{43.78\%} & \textbf{87.63\%} & \textbf{98.34\%} & \textbf{92.68\%} & \textbf{84.08\%} & 734 & \textbf{3} \\
\midrule
\multicolumn{11}{c}{\textbf{French}} \\
\midrule
\multirow{3}{*}{Mixtral 8x7B} & Baseline & 28.00\% & 7.53\% & 11.87\% & 81.72\% & 95.53\% & 88.09\% & 79.01\% & 0 & 0 \\
& CoT prompting & 62.92\% & \textbf{29.68\%} & \textbf{40.33\%} & \textbf{85.54\%} & 95.98\% & 90.46\% & 80.48\% & 0 & 10 \\
& Auto prompt cascade & \textbf{78.12\%} & 26.32\% & 39.37\% & 85.12\% & \textbf{98.28\%} & \textbf{91.23\%} & \textbf{81.59\%} & 793 & \textbf{3} \\
\midrule
\multirow{3}{*}{Mixtral 8x22B} & Baseline & 43.86\% & 13.16\% & 20.25\% & 82.74\% & 96.11\% & 88.93\% & 80.55\% & 0 & 0 \\
& CoT prompting & \textbf{75.00\%} & 25.13\% & 37.65\% & 84.93\% & \textbf{98.05\%} & \textbf{91.02\%} & 81.24\% & 0 & 10 \\
& Auto prompt cascade & 68.97\% & \textbf{31.41\%} & \textbf{43.16\%} & \textbf{85.87\%} & 96.72\% & 90.97\% & \textbf{81.36\%} & 793 & \textbf{3} \\
\bottomrule
\end{tabular}
\caption{Performance comparison of different prompt generation methods on multilingual correctness datasets.}
\label{tab:multilingual_results_full}
\end{table*}

\subsection{English applicability results}
\label{sec:en_applicability}
We also show the full results table of the applicability task (introduced in \S\ref{sec:datasets}). Table~\ref{tab:applicability_results_full} is an expanded version of Table~\ref{tab:applicability_results}, reporting metrics on our English dataset spanning 1,310 unique PC-SA combinations. Across both Mixtral 8x7B and 8x22B, the auto-prompt cascade consistently outperforms baseline and CoT prompting on Inapplicable F1, Applicable F1, and Accuracy. For Mixtral 8x7B, the cascade improves Inapplicable F1 from 51.45\% (baseline) and 52.16\% (CoT) to 56.63\%, Applicable F1 from 78.83\% and 76.84\% to 79.85\%, and Accuracy from 70.52\% to 72.49\%. Similar trends hold for Mixtral 8x22B, where the cascade increases Inapplicable F1 from 51.92\% (baseline) and 53.12\% (CoT) to 54.62\%, Applicable F1 from 82.83\% and 83.91\% to 84.63\%, and Accuracy from 74.70\% to 77.04\%. As with other tasks, the cascade achieves these improvements while scaling to all 1,310 PC-SA combinations with just 3 minutes of human effort per combination, confirming its efficiency and generalizability.

\begin{table*}[t]
\centering
\scriptsize
\setlength{\tabcolsep}{3pt}
\begin{tabular}{@{}ccccccccccc@{}}
\toprule
\textbf{Model} & \textbf{Method} & \multicolumn{3}{c}{\textbf{Inapplicable}} & \multicolumn{3}{c}{\textbf{Applicable}} & \textbf{Accuracy} & \textbf{Unique PC-SA} & \textbf{Effort  } \\
\cmidrule(lr){3-5} \cmidrule(lr){6-8}
 & \textbf{} & \textbf{Precision} & \textbf{Recall} & \textbf{F1 score} & \textbf{Precision} & \textbf{Recall} & \textbf{F1 score} & & \textbf{Count} &  \\
\midrule
\multirow{3}{*}{Mixtral 8x7B} & Baseline & 52\% & 50.93\% & 51.45\% & 78.48\% & \textbf{79.20\%} & 78.83\% & 70.52\% & 0 & 0 \\
 & CoT prompting & 49.27\% & 55.46\% & 52.16\% & 79.13\% & 74.73\% & 76.84\% & 68.82\% & 0 & 10 \\
 & Auto prompt cascade & \textbf{54.83\%} & \textbf{58.56\%} & \textbf{56.63\%} & \textbf{81.09\%} & 78.65\% & \textbf{79.85\%} & \textbf{72.49\%} & 1310 & \textbf{3} \\
\midrule
\multirow{3}{*}{Mixtral 8x22B} & Baseline & 62.25\% & 44.54\% & 51.92\% & 78.20\% & 88.05\% & 82.83\% & 74.70\% & 0 & 0 \\
 & CoT prompting & 66.98\% & 43.92\% & 53.12\% & 78.46\% & 90.42\% & 83.91\% & 76.15\% & 0 & 10 \\
 & Auto prompt cascade & \textbf{69.30\%} & \textbf{45.15\%} & \textbf{54.62\%} & \textbf{78.97\%} & \textbf{91.15\%} & \textbf{84.63\%} & \textbf{77.04\%} & 1310 & \textbf{3} \\
\bottomrule
\end{tabular}
\caption{Performance comparison of different prompt generation methods and models for the applicability task}
\label{tab:applicability_results_full}
\end{table*}

\subsection{Qualitative results}
\label{sec:qualitative_results}

In Table \ref{tab:error_correction}, we present qualitative examples that demonstrate the impact of integrating PC-SA instructions within our auto prompt cascade framework for the correctness task in english. These cases illustrate how targeted, context-aware instructions can substantially improve model performance by guiding the model to attend to PC specific nuances that generic prompts may overlook. By comparing the model’s initial CoT predictions with those informed by PC-SA instructions, we highlight the value of structured, domain-adapted guidance in reducing assessment errors and enhancing interpretability in product quality tasks.

In the first example, for a walking stick's base material, the initial CoT reasoning fails to link a \textit{metal-reinforced removable rubber tip cover} to the SA value  \textit{rubber}. Our auto generated instruction explicitly defines the base material as the component that contacts the ground and provides stability and traction. This PC specific clarification enables the model to correctly infer that the rubber tip, which interacts with the ground, justifies the SA value \textit{rubber} despite the main body being \textit{Huantang wood}. The corrected assessment shows how such context-sensitive instructions can resolve ambiguities arising from indirect evidence in product descriptions.

In the second example, for the \emph{age range} SA of a pet food product, the initial model prediction was correct due to a lack of explicit age information in the product data. However, the PC-SA instruction emphasizes the need to map age range attributes to the intended life stage of the animal, as inferred from product descriptions. This instruction prompted the model to scrutinize the product title and bullet points, which referenced \textit{puppy} and \textit{rapidly growing puppies}, indicating the product targets the puppy growth stage rather than \textit{young adult}. Consequently, the model revised its assessment, identifying the mismatch between the test value and the actual target demographic. This example underscores how PC-SA instructions can surface implicit cues in product data, enabling more accurate and contextually grounded SA validation.

\begin{table*}[t!]
\scriptsize
\begin{tabular}{|p{0.1\textwidth}|p{0.35\textwidth}|p{0.47\textwidth}|}
\hline
\textbf{Product category} & Walking stick & Pet Food \\
\hline
\textbf{Attribute} & base material & age range \\
\hline
\textbf{Attribute value} & rubber & young adult \\
\hline
\textbf{CoT pred.} & Incorrect & Correct \\
\hline
\textbf{CoT reasoning} & The product data mentions that the walking stick is made of 'Huangtang wood' and has a 'steel spike' and 'metal-reinforced rubber tip cover'. There is no mention of the walking stick being made of rubber material. Therefore, the test value 'Rubber' for the attribute 'base material' contradicts the facts stated in the product data. & The product data does not contain any information about the intended age range for this dog food product. The title and description focus on nutritional details for growing puppies, but do not specify an age range. Without clear evidence to confirm or contradict the test value of 'Young Adult', the prediction must be 'Correct'. \\
% \hline 
% % \textbf{Product category Definition} & A walking stick is a device used to facilitate walking or for fashion. It is essentially a long rod that is held by one hand and rests on the ground periodically between steps. Products that are in scope include, but are not limited to: canes and paired hiking poles. & Pet food is a nutritious substance that animals kept as pets eat in order to maintain life and growth. Pet food is usually formulated for a specific type or grouping of animals such as cats, dogs, birds, reptiles, and fish. Products that are in scope include, but are not limited to: primary food or meal, and nutritive treats such as dog biscuits, rawhide bones, or catnip treats. Products that are out of scope include, but are not limited to: pet vitamins or dietary supplements. \\
% % \hline
% % \textbf{Attribute definition} & This attribute provides a description of the material of an item's base. & This attribute indicates the age range that the item is intended or appropriate for. \\
\hline
\textbf{Auto prompt cascade PC-SA instruction} & Base material refers to the material that makes up the bottom part of a walking stick, which comes into contact with the ground and provides stability and traction when using the stick. & Age range indicates the intended or appropriate age group for which the pet food product is suitable. Pet food is designed to provide proper nutrition for animals at different life stages, and the age range helps ensure that the product meets the specific dietary needs of pets based on their age. It guides users to select the most appropriate age range for their pet, ensuring they receive the right nutritional balance for their growth and development stage. \\
\hline
\textbf{Auto prompt cascade pred.} & Correct & Incorrect \\
\hline
\textbf{Auto prompt cascade reasoning} & The product data mentions a 'metal-reinforced removable rubber tip cover' for the walking stick, which implies that the base material that comes into contact with the ground is rubber. Therefore, the test value 'Rubber' for 'base.material' is consistent with the information provided in the product data. & The product title 'Complete And Balanced Dry Dog Food For Growing Puppies 20Lb' and the bullet points mentioning 'puppy' and 'rapidly growing puppies' clearly indicate that this product is intended for puppies or young dogs in their growth stage. The test value 'Young Adult' contradicts this information as it refers to a different life stage beyond the puppy/growing phase. Therefore, the test value is incorrect for this specific product. \\
\hline
\end{tabular}
\caption{Examples of improved correctness decisions with automated PC-SA specific instructions.}
\label{tab:error_correction}
\end{table*}

\subsection{Statistical significance}
\label{sec:statistical_significance}
Table~\ref{tab:stats_eval} reports the statistical significance of the improvements of auto prompt cascade (APC) over baseline, CoT, and human-engineered prompts across multiple models. We perform these tests on our primary task of english correctness. We measure the significance using a paired subsampled bootstrap test with 5,000 class-balanced 80\% draws without replacement and a plus-one correction. 

We compute the $\Delta$ F1 scores and the p-values between APC and each method across all the four models - Mixtral 8x7B, Mixtral 8x22B, DeepSeek R1 Distill Qwen 32B, and Claude 3.5 Sonnet. We observe consistent and significant gains in nearly all comparisons, with p-values below 0.002 in the majority of cases. These results confirm that the performance improvements are robust and not attributable to random variation.

\begin{table}[t]
\centering
\scriptsize
\begin{tabular}{l l c c}

\toprule
\textbf{Model} & \textbf{APC vs method} & \textbf{Mean $\Delta$ F1} & \textbf{p-value} \\
\midrule
\multirow{3}{*}{Mixtral 8$\times$7B} 
& Baseline          & 0.1695 & $<$0.002 \\
& CoT               & 0.0986 & $<$0.002 \\
& Human engineered  & 0.0173 & 0.0353 \\
\midrule
\multirow{3}{*}{Mixtral 8$\times$22B} 
& Baseline          & 0.2503 & $<$0.002 \\
& CoT               & 0.0960 & $<$0.002 \\
& Human engineered  & 0.0556 & $<$0.002 \\
\midrule
\multirow{3}{*}{{\shortstack{DeepSeek R1\\Distill \\ Qwen 32B}}}
& Baseline          & 0.1465 & $<$0.002 \\
& CoT               & 0.0688 & $<$0.002 \\
& Human engineered  & 0.0061 & 0.1615 \\
\midrule
\multirow{3}{*}{Claude 3.5 Sonnet} 
& Baseline          & 0.2377 & $<$0.002 \\
& CoT               & 0.1214 & $<$0.002 \\
& Human engineered  & 0.0762 & $<$0.002 \\
\bottomrule
\end{tabular}
\caption{Statistical evaluation of APC improvements over baseline, CoT, and human-engineered prompts.}

\label{tab:stats_eval}
\end{table}

\end{document}